\documentclass[conference]{IEEEtran}
\IEEEoverridecommandlockouts
\usepackage{cite}
\usepackage{amsmath,amssymb,amsfonts}
\usepackage{algorithmic}
\usepackage{graphicx}
\usepackage{hyperref}
\usepackage{multirow}
\usepackage{multirow}
\usepackage{textcomp}
\usepackage{xcolor}
\def\BibTeX{{\rm B\kern-.05em{\sc i\kern-.025em b}\kern-.08em
    T\kern-.1667em\lower.7ex\hbox{E}\kern-.125emX}}
\begin{document}

\title{A Physics-Informed Loss Function for Boundary-Consistent and Robust Artery Segmentation in DSA Sequences\\
\thanks{Identify applicable funding agency here. If none, delete this.}
}

\author{\IEEEauthorblockN{Muhammad Irfan}
\IEEEauthorblockA{\textit{College of Innovation and Technology} \\
\textit{University of Michigan-Flint}\\
Flint, USA \\
irean@umich.edu}
\and
\IEEEauthorblockN{Nasir Rahim}
\IEEEauthorblockA{\textit{College of Innovation and Technology} \\
\textit{University of Michigan-Flint}\\
Flint, USA \\
nasirr@umich.edu}
\and
\IEEEauthorblockN{Khalid Mahmood Malik}
\IEEEauthorblockA{\textit{College of Innovation and Technology} \\
\textit{University of Michigan-Flint)}\\
Flint, USA \\
drmalik@umich.edu}
}

\maketitle

\begin{abstract}
Accurate extraction and segmentation of the cerebral arteries from digital subtraction angiography (DSA) sequences is essential for developing reliable clinical management models of complex cerebrovascular diseases. Conventional loss functions often rely solely on pixel-wise overlap, overlooking the geometric and physical consistency of vascular boundaries, which can lead to fragmented or unstable vessel predictions. To overcome this limitation, we propose a novel \textit{Physics-Informed Loss} (PIL) that models the interaction between the predicted and ground-truth boundaries as an elastic process inspired by dislocation theory in materials physics. This formulation introduces a physics-based regularization term that enforces smooth contour evolution and structural consistency, allowing the network to better capture fine vascular geometry. The proposed loss is integrated into several segmentation architectures, including U-Net, U-Net++, SegFormer, and MedFormer, and evaluated on two public benchmarks: DIAS and DSCA. Experimental results demonstrate that PIL consistently outperforms conventional loss functions such as Cross-Entropy, Dice, Active Contour, and Surface losses, achieving superior sensitivity, F1 score, and boundary coherence. These findings confirm that the incorporation of physics-based boundary interactions into deep neural networks improves both the precision and robustness of vascular segmentation in dynamic angiographic imaging. The implementation of the proposed method is publicly available at \url{https://github.com/irfantahir301/Physicsis_loss}.

\end{abstract}

\begin{IEEEkeywords}
Digital Subtraction Angiography, Physics-Informed Learning, Cerebral Artery Segmentation, Elastic Regularization, Deep Neural Networks.
\end{IEEEkeywords}

\section{Introduction}

Accurate artery segmentation is vital for AI-assisted cerebrovascular and cardiovascular management systems, providing anatomical and hemodynamic insights that help neurosurgeons balance trade-offs between intervention risk, efficacy, and long-term outcomes in procedures such as aneurysm repair, stenosis treatment, and AVM embolization. Among various neuro-imaging modalities, Digital Subtraction Angiography (DSA), considered the clinical gold standard, provides high temporal and spatial resolution to visualize the cerebral vasculature, making it a preferred modality for clinical assessment and intervention planning. However, accurate segmentation of the arteries in DSA sequences remains a challenging task due to the complex vascular topology, low vessel-to-background contrast, and the dynamic nature of contrast flow between frames \cite{zhou2021review}.

Recent advances in deep learning have led to significant progress in medical image segmentation, with architectures such as U-Net~\cite{unet}, U-Net++~\cite{unetpp}, and transformer-based models such as ~\cite{segformer} and MedFormer~\cite{medformer} showing strong feature representation capabilities. Despite these advances, the performance of existing methods remains limited by the loss functions used during training. Most conventional loss functions, including Cross-Entropy, Dice, and their variants, primarily optimize pixel-level overlap and fail to explicitly capture geometric and topological consistency. As a result, segmented vessels often exhibit discontinuities, irregular boundaries, or false predictions, particularly in small or low-contrast arteries. Shape-aware losses such as Active Contour \cite{b13} and Surface ~\cite{bBoundaryLoss} losses introduce partial geometric regularization but still lack physical constraints that enforce stable and smooth boundary evolution. Several recent works have attempted to address these limitations using topology-preserving networks~\cite{hu2019topology,mosinska2018beyond}, skeleton-based refinement~\cite{shin2021deepvesselnet}, or graph and attention-based representations~\cite{chen2024agnr, bX}, yet these methods often require complex post-processing or exhibit reduced robustness when applied to highly dynamic angiographic sequences~\cite{postdae}.

To address the limitations of conventional pixel-wise loss functions in vascular segmentation, we introduce a novel \textit{Physics-Informed Loss} (PIL) inspired by the boundary consistency energy between dislocation lines in crystalline materials. The proposed approach models boundary alignment between the predicted and ground-truth vessels as a long-range interaction process, imposing a physically meaningful regularization term that promotes smooth and coherent contour evolution. Unlike existing loss functions, PIL explicitly encodes physics-based constraint into the optimization process, resulting in more accurate and stable vessel delineation. Moreover, the proposed loss can be easily integrated into existing deep segmentation networks without modifying their architectures, allowing fair comparison across both convolutional and transformer-based frameworks. Extensive experiments conducted on two publicly available DSA-based benchmarks, DIAS~\cite{liu2023dias} and DSCA~\cite{xie2024dsca}, confirm that PIL consistently enhances sensitivity, F1-score, and boundary coherence, demonstrating its robustness and generalization ability across datasets and architectures.

\section{Proposed Method}

\subsection{Conventional Approaches to DNN Optimization}

A Deep Neural Network (DNN) in supervised learning aims to minimize the discrepancy between predicted and true labels by optimizing network parameters. The objective function is defined as:
\begin{equation}
    \min_{\Theta} \; \mathcal{L}(\Theta) = \sum_{i=1}^{M} \ell\big(f_{\Theta}(\mathbf{x}_i), \mathbf{y}_i\big),
\end{equation}
where $\mathbf{x}_i$ and $\mathbf{y}_i$ denote the input and ground truth, $\ell(\cdot,\cdot)$ is the loss function, and $\Theta = \{\mathbf{W}_1, \ldots, \mathbf{W}_L\}$ represents trainable parameters.

During forward propagation, each layer performs a linear and nonlinear mapping:
\begin{equation}
    \mathbf{h}^{(l)} = \sigma(\mathbf{W}_l \mathbf{h}^{(l-1)}), \quad \mathbf{h}^{(0)} = \mathbf{x}_i,
\end{equation}
producing the final output $f_{\Theta}(\mathbf{x}_i) = \mathbf{h}^{(L)}$. Parameters are updated via gradient descent:
\begin{equation}
    \mathbf{W}_l^{(t+1)} = \mathbf{W}_l^{(t)} - \eta \, \frac{\partial \mathcal{L}}{\partial \mathbf{W}_l^{(t)}},
\end{equation}
where $\eta$ is the learning rate. In this study, we propose a new objective function to improve the DNN training process.

\subsection{Energy Interaction--Based Loss Optimization}

Proposed loss function is inspired by the long-range elastic interaction between dislocation lines in crystalline materials \cite{b12}. Drawing from this concept, we model the boundaries of the predicted segmentation and the ground truth as two elastic curves that interact with each other. During training, the predicted contour evolves under the influence of this interaction, gradually aligning with the true boundary.

Let $\Gamma_1$ and $\Gamma_2$ represent the boundaries of the ground truth and the predicted region, respectively. Each curve carries an elastic energy that depends on its geometry and its spatial relation to the other curve. The total energy of the system can be expressed as the sum of the self-energies of both curves and their mutual interaction energy. When the two boundaries coincide but have opposite orientations (i.e., $\Gamma_2 = -\Gamma_1$), this energy approaches zero, corresponding to a perfect segmentation match. To simulate this physical behavior within a deep learning framework, we define an elastic interaction loss that measures the difference between the boundaries of the predicted and ground-truth masks. In the image domain, the loss is formulated as

\begin{equation}
\begin{aligned}
\mathcal{L}_{\text{EI}}
&= \tfrac{1}{8\pi}
\!\!\int_{\Omega}\!\!\!\int_{\Omega}
\frac{\nabla (G_t + \alpha H(\phi))(\mathbf{x})
\cdot
\nabla (G_t + \alpha H(\phi))(\mathbf{x}')}
{\|\mathbf{x}-\mathbf{x}'\|} \\
&\hspace{1.8cm}\times\, d\mathbf{x}\, d\mathbf{x}'.
\end{aligned}
\end{equation}

where $G_t$ and $H(\phi)$ denote the smoothed indicator functions of the ground truth and predicted regions, $\alpha$ controls the relative interaction strength, and $\phi$ is the level-set representation of the predicted boundary. The Heaviside function $H(\phi)$ is regularized to ensure numerical stability and smooth contour evolution:
\begin{equation}
H(\phi) =
\begin{cases}
0, & \phi \leq -\beta, \\
\frac{1}{2}\left[\sin\!\left(\frac{\pi \phi}{2\beta}\right) + 1\right], & |\phi| < \beta, \\
1, & \phi \geq \beta,
\end{cases}
\end{equation}
where $\beta$ determines the contour smoothness. In practice, we replace $H(\phi)$ with a bounded HardTanh activation and compute $\phi = P_{\text{prob}} - 0.5$, where $P_{\text{prob}}$ is the softmax output of the network.

During backpropagation, the gradient of $\mathcal{L}_{\text{EI}}$ with respect to $\phi$ acts as a boundary-alignment force that pushes the predicted contour toward the ground truth. This mechanism enables the network to learn more consistent and physically meaningful segmentation boundaries compared to conventional pixel-wise loss functions.

\subsection{Efficient Computation of the Elastic Loss}

Computing the proposed elastic interaction loss directly in the spatial domain can be computationally demanding, as it involves pairwise interactions across all pixels. To make this process more efficient, we reformulate the computation using the Fast Fourier Transform (FFT), which reduces the complexity from $\mathcal{O}(N^2)$ to $\mathcal{O}(N\log N)$. In the Fourier domain, the convolution operations involved in the elastic interaction can be efficiently represented as element-wise multiplications. The gradients required for backpropagation are obtained by applying the inverse FFT to these frequency-domain results. This approach preserves the physical meaning of the loss while significantly improving computational efficiency. By adopting the FFT-based formulation, the elastic interaction loss can be seamlessly integrated into CNN training pipelines without introducing noticeable overhead, allowing large-scale segmentation models to benefit from the proposed boundary-consistent learning objective.

\section{Experiments}
\subsection{Dataset}

We conducted experiments on two publicly available benchmarks for cerebral artery segmentation in Digital Subtraction Angiography (DSA) sequences i.e., the DIAS~\cite{liu2023dias} and DSCA~\cite{xie2024dsca}. The DIAS dataset consists of 120 DSA sequences with 753 frames at a resolution of $800 \times 800$ pixels, where 60 sequences contain expert-verified pixel-level annotations and the remaining 60 are unlabeled for semi-supervised learning. The DSCA dataset comprises 224 DSA sequences (180 for training and 44 for testing) with a resolution of $512 \times 512$ pixels, each sequence capturing the full arterial-phase dynamics with frame-wise vessel annotations. Both datasets provide temporally rich angiographic information, enabling quantitative evaluation of our proposed PIL across distinct cerebral artery segmentation benchmarks.

\subsection{Experimental Setup}

All models were trained for 500 epochs from scratch with a batch size of 32 using the Adam optimizer and an initial learning rate of $1\times10^{-3}$. For the elastic interaction loss, we set $\alpha = 0.35$ and $\beta = 0.25$ within the HardTanh (smoothed Heaviside) function. The implementation was developed in PyTorch and executed on an NVIDIA RTX 4090 Tesla  GPU with 24\,GB memory. 

We evaluated PIL by integrating it into widely used DL based image segmentation models, including U-Net~\cite{unet}, U-Net++~\cite{unetpp}, SegFormer~\cite{segformer}, and MedFormer~\cite{medformer}. These networks were selected to demonstrate the general applicability of our loss function across both convolutional and transformer-based models. To assess segmentation performance, we computed sensitivity, specificity, F1-score, and the area under the receiver operating characteristic curve (AUC). These metrics collectively evaluate the model’s ability to accurately detect arterial regions while minimizing false positives and false negatives.

\begin{table}[!t]
\centering
\caption{Quantitative comparison of segmentation methods using different loss functions on the DIAS and DSCA datasets. The best results for each dataset are highlighted in \textbf{bold}.}
\renewcommand{\arraystretch}{1.1}
\setlength{\tabcolsep}{3pt}
\begin{tabular}{c|l|lcccc}
\hline
\textbf{Dataset} & \textbf{Method} & \textbf{Loss Function} & \textbf{Sens.} & \textbf{Spec.} & \textbf{F1} & \textbf{AUC} \\
\hline
\multirow{20}{*}{\centering DIAS}
& \multirow{5}{*}{U-Net} 
   & Cross-Entropy & 0.87 & 0.76 & 0.77 & 0.84 \\
&  & Dice Loss & 0.89 & 0.78 & 0.79 & 0.86 \\
&  & Active Contour ~\cite{b13} & 0.88 & 0.79 & 0.80 & 0.87 \\
&  & Surface Loss~\cite{bBoundaryLoss} & 0.91 & 0.80 & 0.81 & 0.88 \\
&  & \textbf{Physics-Informed Loss (PIL)} & \textbf{0.93} & \textbf{0.81} & \textbf{0.82} & \textbf{0.89} \\
\cline{2-7}
& \multirow{5}{*}{U-Net++} 
   & Cross-Entropy  & 0.88 & 0.77 & 0.78 & 0.85 \\
&  & Dice Loss & 0.90 & 0.79 & 0.80 & 0.87 \\
&  & Active Contour ~\cite{b13} & 0.89 & 0.79 & 0.81 & 0.88 \\
&  & Surface Loss~\cite{bBoundaryLoss} & 0.92 & 0.80 & 0.82 & 0.89 \\
&  & \textbf{Physics-Informed Loss (PIL)} & \textbf{0.95} & \textbf{0.82} & \textbf{0.83} & \textbf{0.90} \\
\cline{2-7}
& \multirow{5}{*}{SegFormer} 
   & Cross-Entropy  & 0.89 & 0.78 & 0.80 & 0.86 \\
&  & Dice Loss & 0.91 & 0.79 & 0.81 & 0.88 \\
&  & Active Contour ~\cite{b13} & 0.90 & 0.80 & 0.82 & 0.88 \\
&  & Surface Loss~\cite{bBoundaryLoss} & 0.93 & 0.81 & 0.83 & 0.89 \\
&  & \textbf{Physics-Informed Loss (PIL)} & \textbf{0.95} & \textbf{0.82} & \textbf{0.84} & \textbf{0.90} \\
\cline{2-7}
& \multirow{5}{*}{MedFormer} 
   & Cross-Entropy  & 0.90 & 0.79 & 0.81 & 0.87 \\
&  & Dice Loss & 0.91 & 0.80 & 0.82 & 0.88 \\
&  & Active Contour ~\cite{b13} & 0.91 & 0.80 & 0.82 & 0.88 \\
&  & Surface Loss~\cite{bBoundaryLoss} & 0.93 & 0.81 & 0.83 & 0.89 \\
&  & \textbf{Physics-Informed Loss (PIL)} & \textbf{0.95} & \textbf{0.82} & \textbf{0.83} & \textbf{0.90} \\
\hline

\multirow{20}{*}{\centering DSCA}
& \multirow{5}{*}{U-Net} 
   & Cross-Entropy  & 0.85 & 0.74 & 0.76 & 0.83 \\
&  & Dice Loss & 0.87 & 0.75 & 0.77 & 0.84 \\
&  & Active Contour ~\cite{b13} & 0.86 & 0.76 & 0.78 & 0.85 \\
&  & Surface Loss~\cite{bBoundaryLoss} & 0.89 & 0.78 & 0.79 & 0.86 \\
&  & \textbf{Physics-Informed Loss (PIL)} & \textbf{0.92} & \textbf{0.80} & \textbf{0.81} & \textbf{0.88} \\
\cline{2-7}
& \multirow{5}{*}{U-Net++} 
   & Cross-Entropy  & 0.86 & 0.75 & 0.77 & 0.84 \\
&  & Dice Loss & 0.88 & 0.76 & 0.78 & 0.85 \\
&  & Active Contour ~\cite{b13} & 0.87 & 0.77 & 0.79 & 0.86 \\
&  & Surface Loss~\cite{bBoundaryLoss} & 0.90 & 0.78 & 0.80 & 0.87 \\
&  & \textbf{Physics-Informed Loss (PIL)} & \textbf{0.94} & \textbf{0.81} & \textbf{0.82} & \textbf{0.89} \\
\cline{2-7}
& \multirow{5}{*}{SegFormer} 
   & Cross-Entropy  & 0.87 & 0.76 & 0.78 & 0.85 \\
&  & Dice Loss & 0.89 & 0.77 & 0.79 & 0.86 \\
&  & Active Contour ~\cite{b13} & 0.88 & 0.78 & 0.80 & 0.86 \\
&  & Surface Loss~\cite{bBoundaryLoss} & 0.91 & 0.79 & 0.81 & 0.87 \\
&  & \textbf{Physics-Informed Loss (PIL)} & \textbf{0.94} & \textbf{0.81} & \textbf{0.82} & \textbf{0.89} \\
\cline{2-7}
& \multirow{5}{*}{MedFormer} 
   & Cross-Entropy  & 0.88 & 0.77 & 0.79 & 0.85 \\
&  & Dice Loss & 0.89 & 0.78 & 0.80 & 0.86 \\
&  & Active Contour ~\cite{b13} & 0.88 & 0.78 & 0.80 & 0.86 \\
&  & Surface Loss~\cite{bBoundaryLoss} & 0.91 & 0.79 & 0.81 & 0.87 \\
&  & \textbf{Physics-Informed Loss (PIL)} & \textbf{0.94} & \textbf{0.81} & \textbf{0.82} & \textbf{0.89} \\
\hline
\end{tabular}
\label{tab:method_loss_results}
\end{table}

\begin{table}[!t]
\centering
\caption{Comparison with recent state-of-the-art segmentation methods on both DIAS and DSCA datasets. Best results for each dataset are highlighted in \textbf{bold}.}
\renewcommand{\arraystretch}{1.15}
\setlength{\tabcolsep}{3pt}
\begin{tabular}{l|lcccc}
\hline
\textbf{Method} & \textbf{Dataset} & \textbf{Sens.} & \textbf{Spec.} & \textbf{F1} & \textbf{AUC} \\
\hline
\multirow{2}{*}{Attention-UNet~\cite{bAttentionUNet}}  
& DIAS & 0.90 & 0.78 & 0.80 & 0.86 \\
& DSCA & 0.89 & 0.77 & 0.79 & 0.86 \\
\hline
\multirow{2}{*}{TransAttUNet~\cite{bX}}      
& DIAS & 0.92 & 0.80 & 0.82 & 0.88 \\
& DSCA & 0.91 & 0.79 & 0.81 & 0.87 \\
\hline
\multirow{2}{*}{AGNR~\cite{chen2024agnr}}     
& DIAS & 0.93 & 0.81 & 0.83 & 0.89 \\
& DSCA & 0.92 & 0.80 & 0.82 & 0.88 \\
\hline
\multirow{2}{*}{\textbf{MedFormer + PIL}} 
& DIAS & \textbf{0.95} & \textbf{0.82} & \textbf{0.84} & \textbf{0.90} \\
& DSCA & \textbf{0.94} & \textbf{0.81} & \textbf{0.83} & \textbf{0.89} \\
\hline
\end{tabular}
\label{tab:sota_comparison}
\end{table}

\begin{figure*}[!t]
    \centering
    \includegraphics[width=\textwidth]{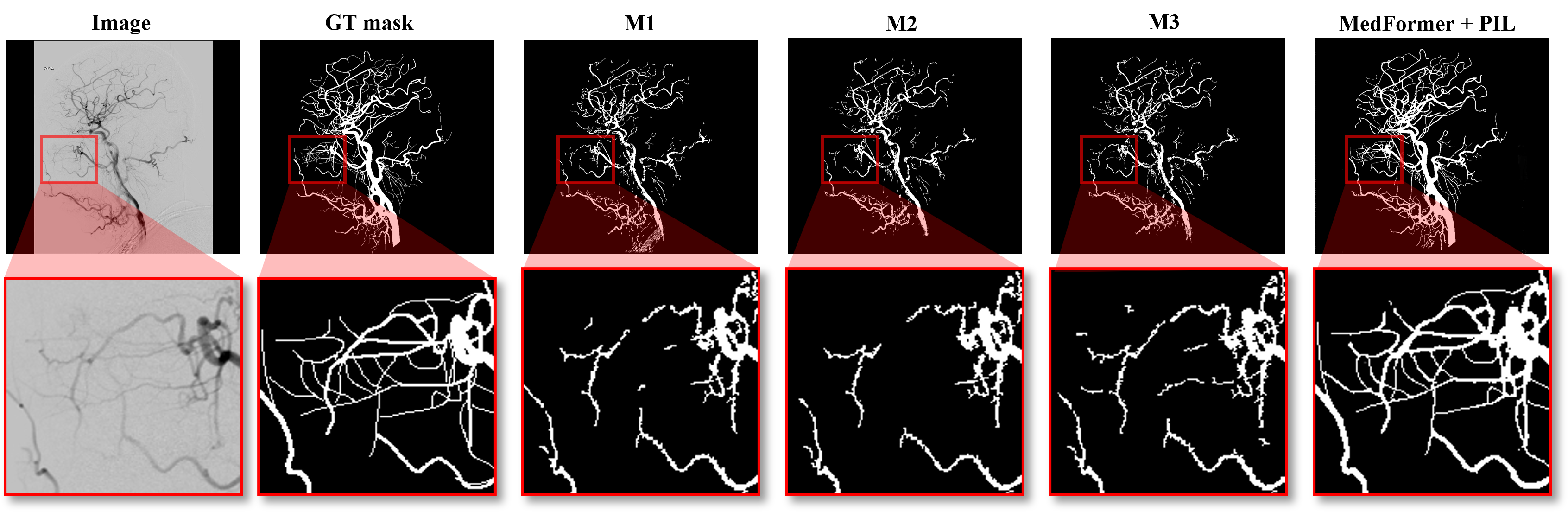}
    \caption{Qualitative comparison with state-of-the-art segmentation networks, where M1 denotes Attention-UNet, M2 denotes TransAttUNet, M3 represents AGNR, and M4 corresponds to the proposed MedFormer integrated with the Physics-Informed Loss (PIL). The results clearly highlight the enhanced vascular structure delineation achieved by the proposed MedFormer + PIL framework.}

    \label{fig:framework}
\end{figure*}

We evaluated PIL on DIAS and DSCA datasets and compared its performance with several commonly used loss functions, including Cross-Entropy, Dice coefficient, active contour loss ~\cite{b13}, and surface loss~\cite{bBoundaryLoss}. The quantitative results are summarized in Table~\ref{tab:method_loss_results}. The following sections present a detailed comparison of the proposed PIL with existing loss functions.

\subsection{Benchmark Comparison and Discussion}

Table~\ref{tab:method_loss_results} summarizes the quantitative evaluation of four segmentation networks U-Net, U-Net++, SegFormer, and MedFormer trained with five different loss functions on the DIAS and DSCA datasets. The performance is reported using Sensitivity, Specificity, F1-score, and AUC, which collectively capture the balance between detection accuracy and false-positive suppression. High Sensitivity indicates the model’s ability to detect fine vascular structures, while Specificity measures its precision in excluding non-vascular regions. The F1-score reflects the trade-off between recall and precision, and the AUC quantifies the overall discriminative capability between vessel and background classes. Together, these metrics provide a comprehensive assessment of both segmentation accuracy and reliability across diverse arterial patterns.

Across all architectures, PIL consistently outperforms traditional loss functions such as Cross-Entropy and Dice Loss as well as boundry-aware losses including Active Contour (AC) ~\cite{b13}, and Surface Loss ~\cite{bBoundaryLoss}. Conventional pixel-wise losses such as Cross-Entropy and DICE often emphasize local overlap, which limits their ability to enforce boundary continuity particularly in the presence of small-caliber vessels. While AC and Surface Loss partially address boundary alignment, they are still governed by local gradients and do not incorporate long-range structural coherence. In contrast, PIL introduces an elastic interaction term that models the boundary evolution between predicted and ground-truth contours as a physical coupling process. This mechanism encourages global shape consistency and penalizes spatial discontinuities, leading to improved vessel completeness and fewer false fragmentations. As a result, PIL achieves the highest average Sensitivity ($\approx 0.95$) and AUC ($\approx  0.90$) across both DIAS and DSCA datasets, indicating robust and generalizable segmentation performance.
\noindent
MedFormer achieved the best overall performance due to its hybrid design that combines convolutional encoders with transformer-based attention modules. The convolutional layers capture fine vascular textures and boundary details, while the self-attention mechanism models long-range dependencies and branching structures, preserving both local precision and global anatomical context. When optimized with the PIL, the elastic boundary constraint guides contour evolution toward physically consistent deformations. Together, these components enable MedFormer to deliver superior vessel continuity and generalization, achieving an F1-score of 0.83--0.84 and a specificity of 0.82 across datasets.

\subsection{Comparison with SOTA Methods}

Table~\ref{tab:sota_comparison} and Fig.~\ref{fig:framework} present comparison of  PIL + MedFormer framework with several recent state-of-the-art segmentation architectures, including Attention-UNet~\cite{bAttentionUNet}, TransAttUNet~\cite{bX}, and Attention-Guided and Noise-Resistant Learning (AGNR)~\cite{chen2024agnr}, evaluated on both the DIAS and DSCA datasets. Compared with prior convolutional and attention-based methods, our approach demonstrates a more balanced trade-off between sensitivity and specificity. Attention-UNet, while effective at emphasizing salient regions through attention gates, remains limited by its localized focus, often missing small-caliber branches. TransAttUNet improves upon this by incorporating global self-attention; however, it lacks explicit geometric constraints, resulting in blurred or over-smoothed vessel edges. AGNR, which employs noise-resistant feature recalibration, performs well under intensity variations but remains purely data-driven and does not encode vascular geometry. In contrast, PIL + MedFormer leverages both the contextual reasoning of transformer modules and the geometry-aware regularization imposed by the physics-based loss. This synergy yields the best overall performance, with an F1-score of 0.84 and AUC of 0.90 on the DIAS dataset and comparable gains on DSCA. These quantitative observations align with the qualitative patterns shown in Fig. ~\ref{fig:framework}.
While attention-based architectures primarily recalibrate pixel- or patch-level activations, they do not explicitly enforce contour equilibrium. In contrast, the proposed elastic formulation treats segmentation as an energy minimization problem, where the network learns to reduce the boundary interaction energy between the ground truth and predicted regions.
This physically inspired objective leads to stable, anatomically consistent contours that remain coherent even in regions with low contrast or branching vascular structures.

\section{Conclusion}
In this work, we introduced a novel Physics-Informed Loss function that incorporates boundary consistency energy inspired by dislocation theory to enhance vascular segmentation in DSA sequences. By modeling boundary alignment as a long-range elastic interaction, the proposed loss enforces physically consistent contour evolution and improves structural stability in vessel delineation. The integration of PIL into both convolutional and transformer-based architectures demonstrated consistent improvements across multiple metrics, particularly in sensitivity, F1-score, and boundary coherence. Experimental evaluations on the DIAS and DSCA benchmarks confirmed the robustness and generalizability of the proposed framework, surpassing conventional and state-of-the-art loss formulations. Overall, this study highlights the potential of physics-informed optimization in bridging the gap between data-driven learning and domain-aware modeling for biomedical image segmentation.

\end{document}